\documentclass[letterpaper, 10 pt, conference]{ieeeconf}  
\usepackage{graphicx}
\usepackage{cite}
\usepackage{color}
\usepackage{fancyhdr}

\IEEEoverridecommandlockouts                              

\begin{document}

\title{\LARGE \bf
WHERE-Bot: a Wheel-less Helical-ring Everting Robot Capable of Omnidirectional Locomotion
}

\author{Siyuan Feng$^{1}$, Dengfeng Yan$^{1}$, Jin Liu$^{1}$, Haotong Han$^{2}$, Alexandra Kühl$^{1,3}$, and Shuguang Li$^{1,4}$
\thanks{$^{1}$Siyuan Feng, Dengfeng Yan, Jin Liu, Alexandra Kühl, and Shuguang Li are with the Department of Mechanical Engineering, Tsinghua University, Beijing, 100084, China (e-mail: lisglab@tsinghua.edu.cn).}
\thanks{$^{2}$Haotong Han is with the Department of Mechanical Engineering, Stanford University, CA 94305, USA.}
\thanks{$^{3}$Alexandra Kühl is also with the Faculty of Mechanical Engineering, RWTH Aachen University, Aachen, 52062, Germany.}%
\thanks{$^{4}$Shuguang Li is also with Beijing Key Laboratory of Transformative High-end Manufacturing Equipment and Technology, Tsinghua University, Beijing 100084, China.}%
}

\maketitle
\thispagestyle{fancy}
\pagestyle{empty}

\lhead{}
\lfoot{}
\cfoot{\small{Copyright \copyright 2025 IEEE. Personal use of this material is permitted. \\
		However, permission to use this material for any other purposes must be obtained 
		from the IEEE\\by sending an email to pubs-permissions@ieee.org.}}
\rfoot{}
\renewcommand{\headrulewidth}{0mm}

\begin{abstract}


Compared to conventional wheeled transportation systems designed for flat surfaces, soft robots exhibit exceptional adaptability to various terrains, enabling stable movement in complex environments. However, due to the risk of collision with obstacles and barriers, most soft robots rely on sensors for navigation in unstructured environments with uncertain boundaries. In this work, we present the WHERE-Bot, a wheel-less everting soft robot capable of omnidirectional
locomotion. Our WHERE-Bot can navigate through unstructured environments by leveraging its structural and motion advantages rather than relying on sensors for boundary detection. By configuring a spring toy ``Slinky'' into a loop shape, the WHERE-Bot performs multiple rotational motions: spiral-rotating along the hub circumference, self-rotating around the hub's center, and orbiting around a certain point. The robot's trajectories can be reprogrammed by actively altering its mass distribution. The WHERE-Bot shows significant potential for boundary exploration in unstructured environments.

\end{abstract}

\section{INTRODUCTION}

\bibliographystyle{IEEEtran} 
The wheel is one of the most important mechanisms for robot locomotion, characterized by its simple structure and high energy efficiency\cite{rubio2019review,lee2024variable}. While mobile robots or transportation systems equipped with wheels can move fast on flat surfaces, their travel speed diminishes on complex terrains such as sand, grass, mud, and rugged land. Additionally, obstacles significantly higher than the wheel size can obstruct the robot's movement. In contrast to wheeled systems optimized for flat ground travel, soft robots exhibit remarkable adaptability to diverse terrains, allowing them to navigate complex environments\cite{rus2015design}. 

Leveraging soft robotics principles to design wheels can improve environmental adaptability while maintaining high velocity and energy efficiency. Soft actuators enable the construction of deformable wheels that can actively adjust their size according to the terrains encountered\cite{lee2013deformable}. This configuration variability allows the wheel to overcome obstacles higher than its initial size and access narrow gaps lower than its original diameter. Wheels designed with soft surfaces can endure substantial deformation without incurring permanent damage to the structure\cite{lee2024variable,zambelli2021research}. Wheels built entirely from soft materials can even withstand falling from heights significantly greater than the wheel size\cite{gong2016rotary}, which presents the potential for robots to move down cliffs. Additionally, wheels designed with origami patterns are lightweight while offering a relatively high payload-to-weight ratio\cite{lee2017origami}. The transmission ratio between motor torque and ground reaction force can be actively adjusted through origami folding and unfolding, allowing the system to adapt to varying payloads and ground friction conditions\cite{felton2014passive}.

Despite their exceptional adaptability, most soft robots still rely on sensors and controllers to navigate unstructured environments with unknown boundaries. When sensor data is inaccurate due to signal interference or sensor failures, robots are prone to collisions with boundaries or obstacles, particularly in confined spaces. These collisions can hinder locomotion or even damage the robot's structure.
\begin{figure}
    \centering

    \includegraphics[width=0.48\textwidth]{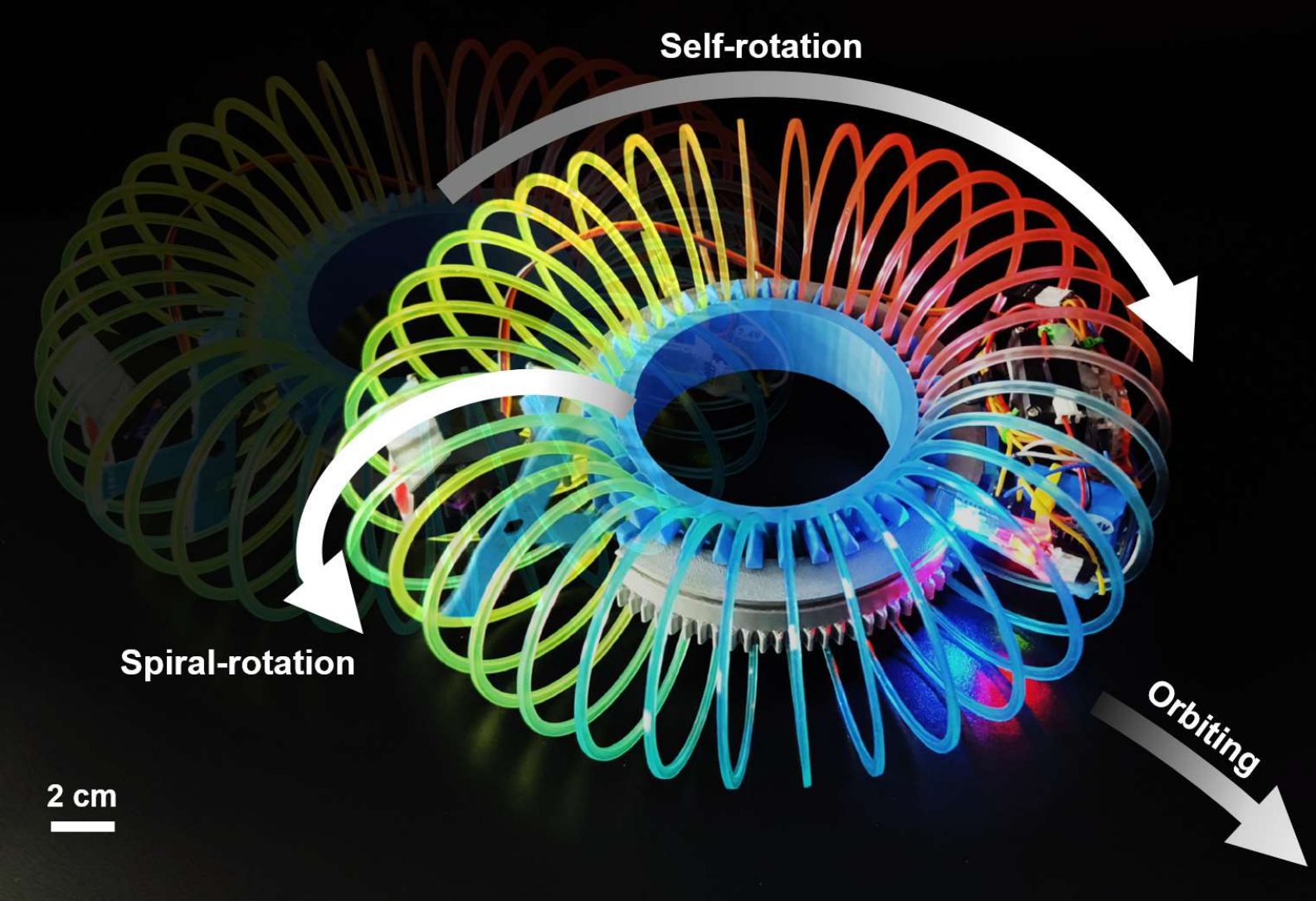}
    \caption{The WHERE-Bot's motion consists of a spiral-rotating (everting) along the hub circumference, a self-rotating around the hub's center, and an orbiting around a certain point.}

    \label{figSHAPE}

\end{figure}

Soft robots utilizing everting motion (turning inside-out) demonstrate strong environmental interaction capabilities\cite{eken2023continuous,hong2009whole,orekhov2010actuation,orekhov2010mechanics,watanabe2022toroidal,coad2019vine,hawkes2017soft,sui2022bioinspired,perez2022self,qi2024defected}. An elongated toroid robot working by turning itself inside out in a single continuous motion is allowed to traverse a complex terrain\cite{hong2009whole}. A soft pneumatic robot that employs everting motion for growth can navigate tightly constrained spaces while forming useful three-dimensional structures along its path\cite{hawkes2017soft}. Based on the everting motion, a soft gripper designed for universal grasping can envelop various objects, achieving a flexible and passive form-fitting grasp\cite{sui2022bioinspired}. An everting toroidal robot propelled by a motorized mechanism exhibits good environmental adaptability, successfully navigating mazes and climbing pipes\cite{perez2022self}. An untethered soft-everting robot is capable of locomoting on sand as well as digging in sand by turning itself inside out\cite{eken2023continuous}. A thermally-actuated liquid crystal elastomer (LCE) soft robot with an asymmetric defect employs a flipping motion, which is also turning inside-out, to achieve periodic spin-orbiting behaviors. This robot has the potential for intelligently mapping the geometric boundaries of unknown confined spaces\cite{qi2024defected}.

Inspired by a spring toy's everting motion, we propose a Wheel-less HElical-Ring Everting mobile robot (the WHERE-Bot, Fig.~\ref{figSHAPE}), which can move in unstructured environments by utilizing its structural and motion advantages to interact with environments. Our robot features a ring-shaped helical structure that facilitates multiple rotational motions during locomotion. With compact structure and flexible steering, our robot can actively change its mass distribution to program trajectories, presenting significant potential for boundary exploration in unstructured environments. 

The rest of this paper is structured as follows: Section II covers the principle of motion, along with the design and fabrication of the robot. In Section III, we examined the robot's basic motions via experiments. In Section IV, we introduce a simplified model to describe the robot's behaviors. In Section V, we present the measurements of the robot's motion parameters, validate the model by comparing predicted and observed trajectories, and demonstrate trajectory control by adjusting its mass distribution manually. We also showcase some potential applications for boundary exploration in unstructured environments. Finally, we discuss and conclude the work in Section VI.

\section{Design and Fabrication }

\begin{figure}
    \centering

    \includegraphics[width=0.48\textwidth]{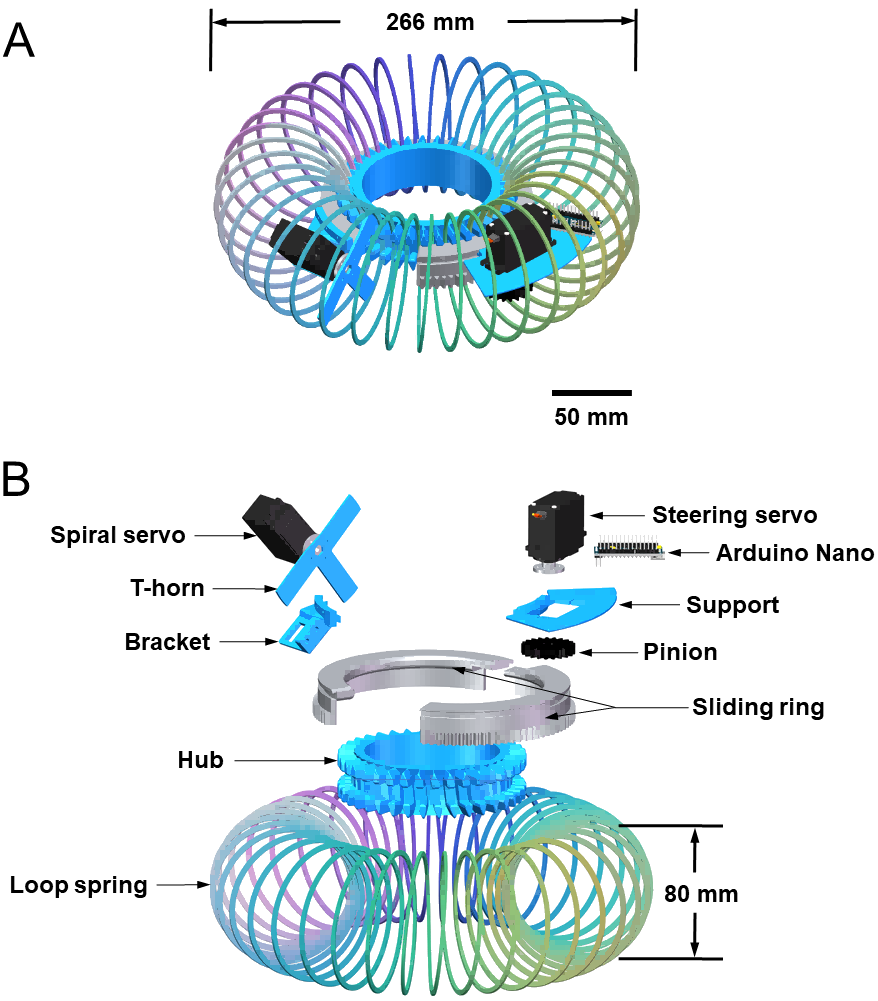}
    \caption{Structure of the WHERE-Bot. (A) Assembled WHERE-Bot. (B) Components of the robot.}

    \label{fig:2}

\end{figure}

The WHERE-Bot is an electrically-driven periodic robot that features the helix-ring body design, which is inspired by the everting motion of a spiral structure. The robot is mainly composed of a helix-looped body prepared by the toy spring ``Slinky", a flexible T-horn, and circuit components. The total weight of the robot is 436 grams. The nominal diameter of the loop is 266 mm (Fig.~\ref{fig:2}(A)), and the total height of the robot is 80 mm (Fig.~\ref{fig:2}(B)). The mechanical section includes a loop spring consisting of 37 coils, a hub arranged in the center, and a sliding ring. The electrical components consist of a circuit board, a Bluetooth module, and two servos for the actuation of spiral-rotation and steering rotation.

\subsection{Principle of Motions}

Unlike conventional wheels, the locomotion of the WHERE-Bot features three motions: the everting of the coils, the self-rotation of the robot, and the orbiting motion, which can be attributed to its unique structural composition. A loop spring is constrained on the wheel hub, and each coil of the spring meshes with the helical teeth on the hub, forming a worm-gear pair. When a torque is applied to a certain spring coil, this coil rotates. Since the spring is a continuous body, this rotation is transmitted to all spring coils, causing each coil to evert. Due to the meshing, the extrusion force exerted by the spring coils on the helical teeth makes the hub rotate.

Under actual locomotion conditions, friction enables the robot to move in a particular direction. The WHERE-Bot has 36 spring coils in contact with the ground, and the frictional forces acting on each coil are different. Their resultant force changes the moving direction of the robot. It can be observed that the robot usually moves in the direction away from the center-of-gravity distribution, suggesting that the center-of-gravity distribution of the robot's structure is the main factor affecting the magnitude of the frictional force. The spring coils closer to the center of gravity are subjected to a greater frictional force. During the movement, the hub rotates, causing the center of gravity to rotate. Thus, the direction of the resultant force acting on the robot also rotates regularly during the movement, enabling the robot to move in an orbit.

\subsection{Structure of the Robot}

The loop spring and the hub are incorporated into the structural design to establish the intended engagement and motion pattern. The loop spring, constructed by bonding the ends of a toy spring known as ``Slinky", is situated within the hub, which is designed as a helical gear. This gear features a pitch circle diameter of 112 mm and an outside diameter of 125 mm, with the spring ring constrained within the gear's bottom land. The helical gear’s helix angle (5 degrees) was empirically selected to enable the continuous eversion and rotation of the spring coil.

A sliding ring is introduced to prevent the loop spring from disengaging from the grooves during locomotion. It stabilizes the innermost section of each spring coil against the groove’s base within the hub. This is achieved by inwardly cutting the hub teeth to a diameter of 105 mm to accommodate the sliding ring with an inner diameter of 108 mm. A horizontal groove is incorporated into the sliding ring, ensuring that the electronics (steering servo, MCU, and battery) on the support module can move smoothly around the ring.

In the circuit module design, two servos are employed to provide driving force and steering function. The spiral servo is mounted to the sliding ring using a 3D-printed bracket, with its rotor concentrically affixed to a pre-designed T-shaped flexible servo horn. This configuration transmits the servo's rotation through the T-horn to the attached spring coil. To maintain continuous locomotion, the T-horn is designed with sufficient flexibility and elasticity to accommodate the deformation of the spring during operation, which might otherwise alter the T-horn's orientation, potentially causing a collision with the hub teeth and impeding movement. The steering servo, along with additional circuit components, is mounted within the groove of the sliding ring using a separate support. In subsequent experiments, we aim to validate whether mass distribution could effectively control the robot's moving direction. To this end, the bottom of the sliding ring is designed as a curved rack, engaging with a pinion connected to the steering servo's rotor. By actuating the steering servo and engaging the rack, the support's position on the sliding ring can be adjusted, thereby modifying the mass distribution and consequently influencing the robot's directional movement.

\begin{figure}
    \centering

    \includegraphics[width=0.48\textwidth]{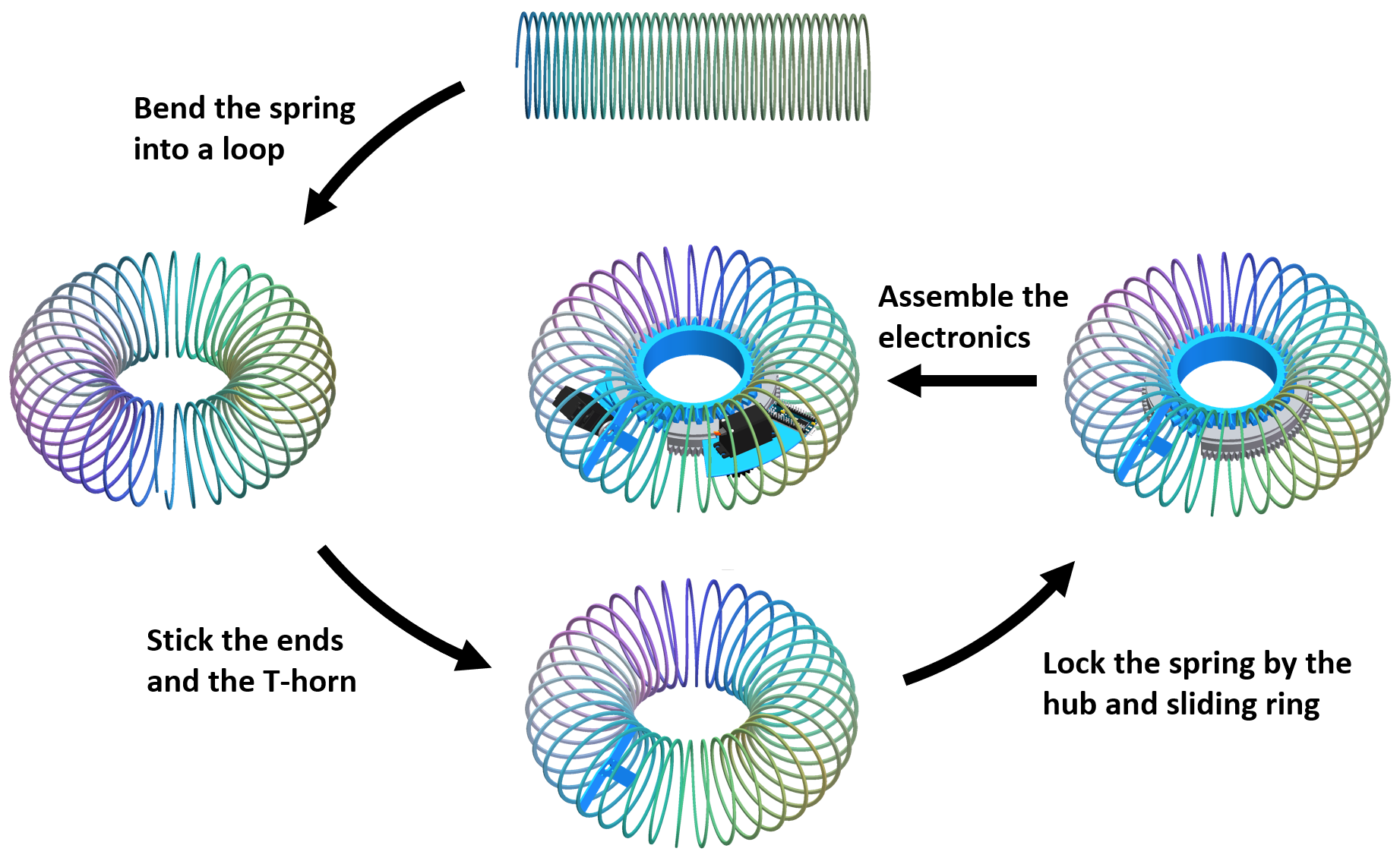}
    \caption{Fabrication of a WHERE-Bot. The spring is a ``Slinky" toy, and other components, excluding servos and circuits, are 3D-printed.}

    \label{fig:3}

\end{figure}

\subsection{Fabrication Method and Material Choice}

As shown in Fig.~\ref{fig:3}, the fabrication process of a WHERE-Bot consists of the following steps: the loop spring construction, the robot body preparation, and the electronics assembly. Materials for the loop spring need to meet the requirements of flexibility, elasticity, and toughness. The ``Slinky" is adopted as it exhibits good resistance to large deformation during transmission. The spring is cut to a specific number of turns and stuck end to end to realize the loop design. The rest of the parts can be 3D-printed, including the hub, the sliding ring, and two sliding carriers (the bracket for the spiral servo and the support for the circuit and the steering servo). The ideal materials for these parts have to be lightweight, so polylactic acid (PLA) is selected for their preparation. Thermoplastic polyurethane (TPU) is selected as the material for building the flexible T-horn. After the parts are prepared, the loop spring is wrapped around the center hub, with each coil restrained in the tooth grooves of the hub by the outer sliding ring. The T-horn is fixed to a spring coil so that the spiral servo can rotate this coil. The Carriers
are finally mounted to the ring. The support for the circuit and steering servo is not fully constrained to the ring but has a degree of freedom to slide around the sliding ring, like a planet gear.

\section{Multiple rotational motions}

The WHERE-Bot was tested to analyze its motions and behaviors. It was found that the loop spring deformed when rotated by the spiral servo, leading to a process of elastic potential energy accumulation and release during the WHERE-Bot's motions. Additionally, the relationship between the relative motions of the loop spring and the hub is analyzed. 

\subsection{Spiral-rotation (Everting Motion)} 
Driven by the spiral servo, the loop spring performs a spiral motion along the circumferential direction of the hub, turning inside out continuously. We refer to this everting motion as spiral-rotation. As shown in Fig.~\ref{fig:5}(A), upon the initiation of the spiral servo, a coil of the loop spring connected to the T-horn (referred to as the driving coil) is subjected to a gradually increasing torque (referred to as the driving torque), increasing friction between the driving coil and the contact surface. When the driving force exceeds the maximum static friction, the driving coil's state will change from rest to rotation while other coils remain at rest. Due to the relative motion of the driving coil and its adjacent coils, the stress (transmitting the driving torque to other coils) in the loop spring increases. When the stress exceeds the friction resistance, other coils begin to rotate.

\begin{figure}
    \centering

    \includegraphics[width=0.48\textwidth]{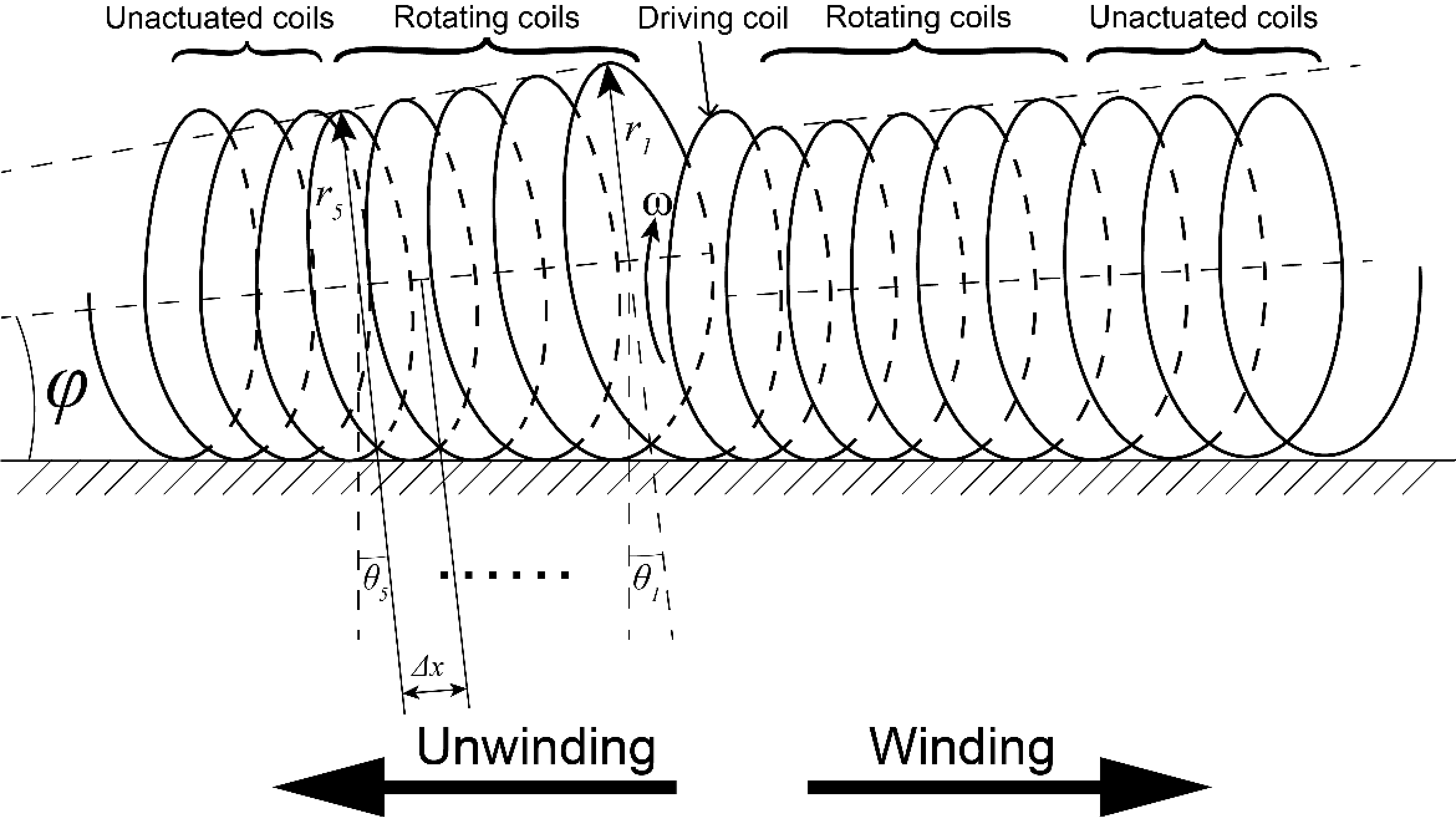}
    \caption{Principle of the spiral-rotation. The axis of the driving coil and some of its adjacent coils transform from a loop into a line, shown in a side view.}

    \label{fig:4}

\end{figure}

As shown in Fig.~\ref{fig:4}, the driving coil rotates at a constant speed, and part of the other coils initiate rotation (rotating coils), while the rest portion remains stationary (unactuated coils). The rotating coils located on the left (right) of the driving coil are in the direction of unwinding (winding) rotation. The coils situated in the direction of unwinding (winding) rotation undergo expansion (contraction). The degree of expansion or contraction depends on the proximity of the coil to the driving coil. The closer the coil is to the driving coil, the more it deforms. This process is defined as the elastic potential energy accumulation phase, during which the driving coil rotates, and the robot remains stationary.

After all the coils transition from a static to a rotating state, the accumulated elastic potential energy will be released rapidly. All the coils, except the driving coil, undergo an accelerating rotation. The robot's movement can be defined as the elastic potential energy release phase.

During the elastic potential energy release phase, the robot attains its maximum moving speed when the stress is released to a critical point where the transmitted driving torque equals the resisting torque due to kinetic friction force. After the friction dissipates the kinetic energy, the robot advances a specific distance and returns to the initial state. After that, the robot continues cycling through the elastic potential energy accumulation and release phases, propelling itself forward.

\subsection{Orbiting}

The trajectory over a single cycle can be discretized into a segment of arcs and a segment of straight lines. When this trajectory is repeated multiple times, it assembles a circle (Fig.~\ref{fig:5}(B)). The angle between the spiral servo and the steering module is modified by regulating the position of the steering module on the sliding ring (Fig.~\ref{fig:5}(D)), thereby altering the orbit radius in real time.

\subsection{Self-rotation}

\begin{figure}
    \centering

    \includegraphics[width=0.48\textwidth]{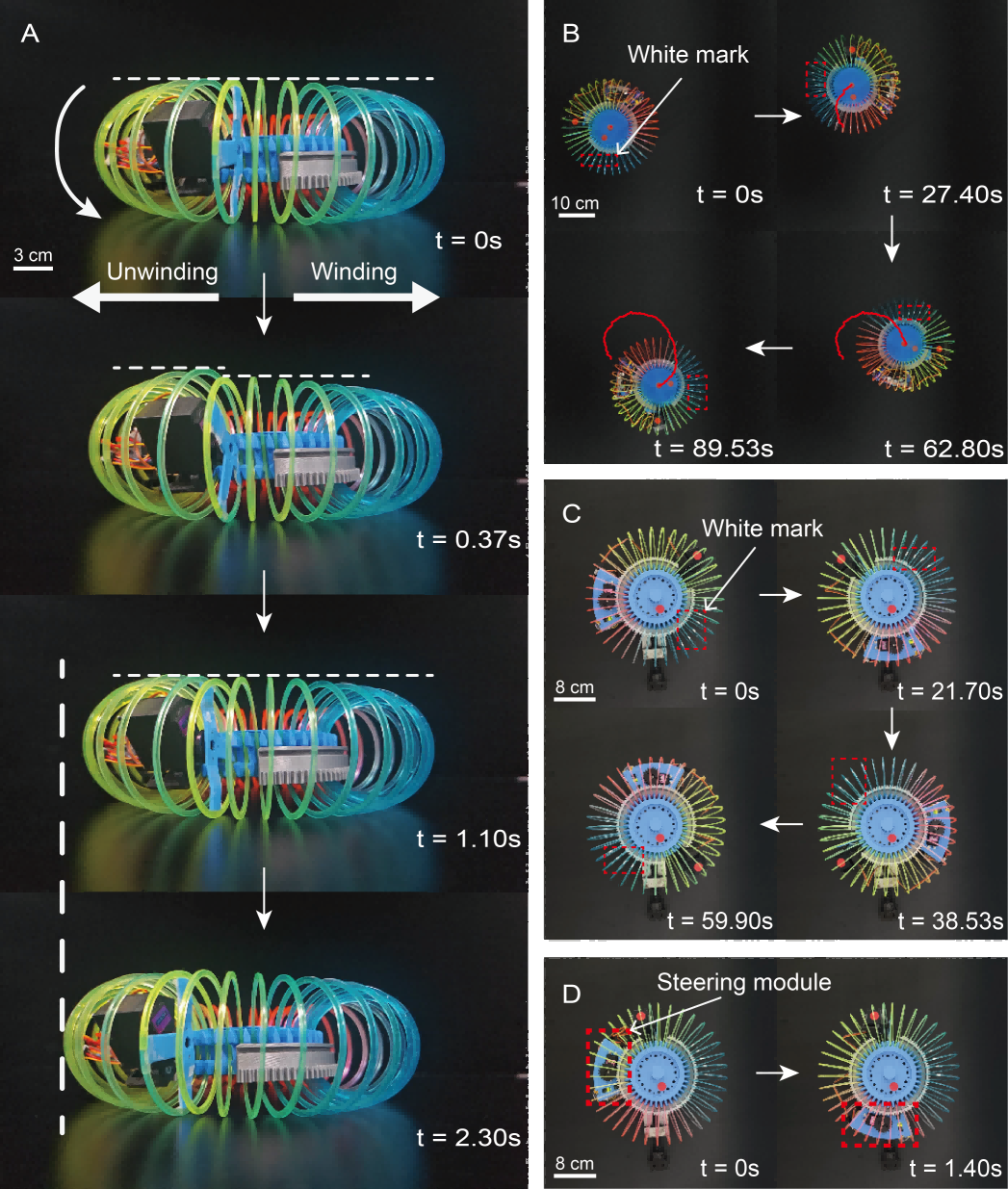}
    \caption{
        Demonstration of multiple rotation motions of the WHERE-Bot. 
        (A) Spiral-rotation (everting motion). 
        (B) Orbiting. 
        (C) Self-rotation. 
        (D) Steering module rotation. 
        }

    \label{fig:5}

\end{figure}

The hub incorporates a groove for each coil in the loop spring, with the grooves evenly spaced around the circumference and separated by gear tooth. As shown in Fig.~\ref{fig:5}(C), when the loop spring is in motion, the hub and the loop spring are similar to a pair of worm gear. When the spiral servo actuates the driving coil to rotate a circle, the loop spring rotates an angle of a hub tooth (Fig.~\ref{fig:5}(A)).

\section{Modeling}

Through our preliminary experiments, we observed that elastic potential energy accumulation and release repeatedly happened in the robot's motions. Consequently, the robot periodically rotates at a certain angle and moves forward a specific distance. To explain how mass distribution affects the robot's moving direction and trajectory, we develop a simplified model of the robot's motions.
\begin{figure}
    \centering

    \includegraphics[width=0.48\textwidth]{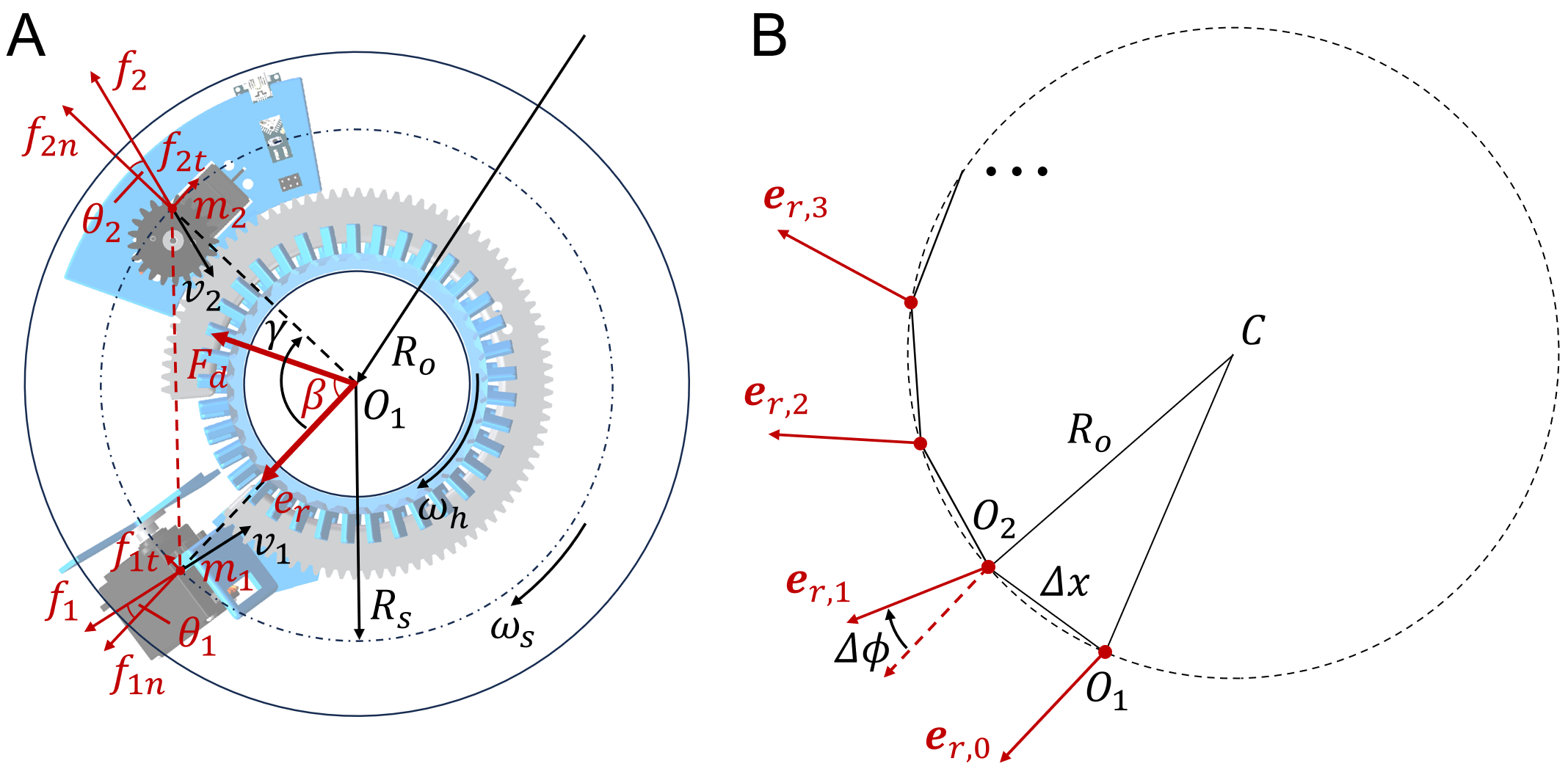}
    \caption{Modeling of the WHERE-Bot. (A) Kinematic analysis diagram presented on the bottom view. (B) Discretized trajectory of the simplified periodic motion.}

    \label{figModeling}

\end{figure}
Before incorporating the spiral servo and steering servo, the robot's structure exhibits an approximately uniform mass distribution, resulting in identical sliding friction magnitudes at the spring's contact points with the ground. The friction at all the contact points can be decomposed into radial and tangential components. The radial components are balanced, while the tangential components generate a friction torque $M_0$ at the robot's center. The mass of the initial structure is denoted as $m_0$. 

Upon adding servos, represented by concentrated masses $m_1$ and $m_2$, on the sliding ring, as shown in Fig.~\ref{figModeling}(A), there is a significant increase in pressure on the spring coils under each servo. This leads to increases in sliding friction, denoted as $f_1,f_2$. The line connecting the robot's center to concentrated mass $m_1$ is defined as the reference direction of the robot.

The angle between the tangential directions of the spring coil and the radial direction at the ground contact point, $\theta_1$ and $\theta_2$, change due to the non-linear elastic deformation of the springs. The tangential components of $f_1,f_2$ contribute to a friction torque increase:
\begin{equation}\label{eqn-1} 
  \Delta M=R_sf_1\sin\theta_1+R_sf_2\sin\theta_2,
\end{equation}
where $R_s$ is the distance from the spring's contact point with the ground to the robot's center. The radial components form a driving force:
\begin{equation}\label{eqn-2} 
F_d=f_1\cos\theta_1\cos\beta+f_2\cos\theta_2\cos(\gamma-\beta),
\end{equation}
where $\beta$ is the angle between the direction of $F_d$ and the reference direction of the robot, and $\gamma$ is the angle between lines connecting the robot's center to concentrated masses $m_1$ and $m_2$. By applying vector composition and inner product calculations, $\beta$ can be determined as follows:

\begin{equation}\label{eqn-3} 
\cos\beta=\frac{(f_{1n}^2+f_{1n}f_{2n}\cos\gamma)}{f_{1n}\sqrt{f_{1n}^2+f_{2n}^2+2f_{1n}f_{2n}\cos\gamma}},
\end{equation}
where $f_{1n}$ and $f_{2n}$ are the radial components of $f_1$ and $f_2$.

At the moment the robot starts moving from a static state, $F_d$ drives the robot forward, while $M_0+\Delta M$ causes it to rotate. The initial linear and angular accelerations can be calculated as:
\begin{equation}\label{eqn-4} 
a_0=\frac{F_d}{m_0+m_1+m_2},
\end{equation}
\begin{equation}\label{eqn-5} 
\alpha_0=\frac{M_0+\Delta M}{J},
\end{equation}
where $J$ represents the robot's moment of inertia.

When the robot moves on a flat surface with isotropic friction, $a_0$ and $\alpha_0$ determine the trajectory during each period. Given the values of size, mass, and inertia, $a_0$ and $\alpha_0$ depend solely on $\gamma$. Thus, $\gamma$ determines the direction and distance of the robot's periodic motion.

As shown in Fig.~\ref{figModeling}(B), the robot's periodic movements, characterized by a constant turning angle $\Delta\phi$ and moving distance $\Delta x$, make the positions between periods located on a circle with radius $R_o$.

Assuming the robot completes one orbit in $n$ periods ($n$ may not always be an integer due to uncertain divisibility of 360 degrees by $\Delta\phi$), $R_o$ can be estimated as:
\begin{equation}\label{eqn-6} 
R_o\approx\frac{1}{2}\sum_{k=1}^{\lfloor n/2\rfloor}\Delta x\cdot\sin(k\Delta\phi).
\end{equation}

This model allows us to predict the robot's moving direction and trajectory based on the angular position of the steering servo.

\section{Locomotion Experiments}

Our robot features motions pre-defined by mass distribution, reprogrammable trajectory, and the ability to explore the environment autonomously. We conducted a series of experiments to demonstrate these capabilities. During the experiments, markers were affixed to the robot's components (a cover was 3D-printed and fixed on the top of the hub for marker attachment). We recorded the movement process at 60 fps using a camera and utilized software \textit{Tracker} to obtain the positions of the markers.
\subsection{Motions Pre-defined by the Initial Mass Distribution}

The robot was placed on a hard and flat table to move with a fixed $\gamma$ during its motion. $\gamma$ was set from 60 to 300 degrees, with an increment of 60 degrees. For each value of $\gamma$, three experiments were conducted. We processed the trajectory using least-squares circle fitting and obtained the fitted orbit radius. The periodic turning angle $\Delta\phi$ was calculated by dividing the total directional change of the robot by the number of motion periods. Subsequently, we estimated the periodic moving distance $\Delta x$ using Equation \ref{eqn-6}. The motion period $T$ was determined by dividing the total motion time by the number of periods, which was counted manually. The moving direction $\beta$ was then found by calculating the inner product of the moving direction in each period and the reference direction at the beginning of each period.
\begin{figure}
    \centering

    \includegraphics[width=0.48\textwidth]{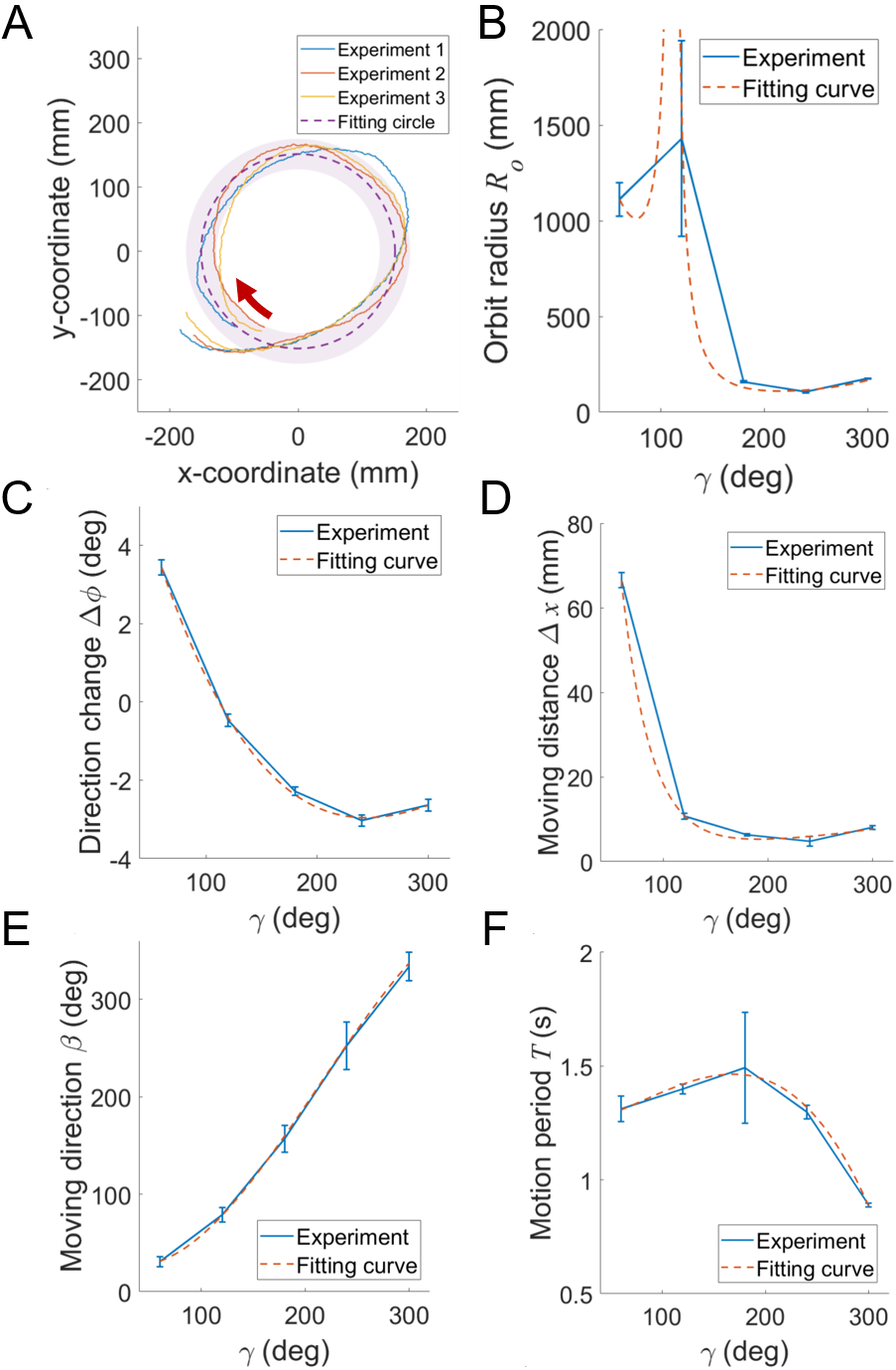}
    \caption{Trajectory programmed by mass distribution and motion parameters as a function of structural parameter $\gamma$. (A) Actual trajectory and fitting circle at $\gamma=180\rm\,deg$. (B-F) We recorded (B) orbit radius, (C) direction change, (D) moving distance, (E) moving direction, and (F) motion period as a function of $\gamma$. Common functions were used to fit the relations.}

    \label{figResults}

\end{figure}

\begin{figure}
    \centering

    \includegraphics[width=0.48\textwidth]{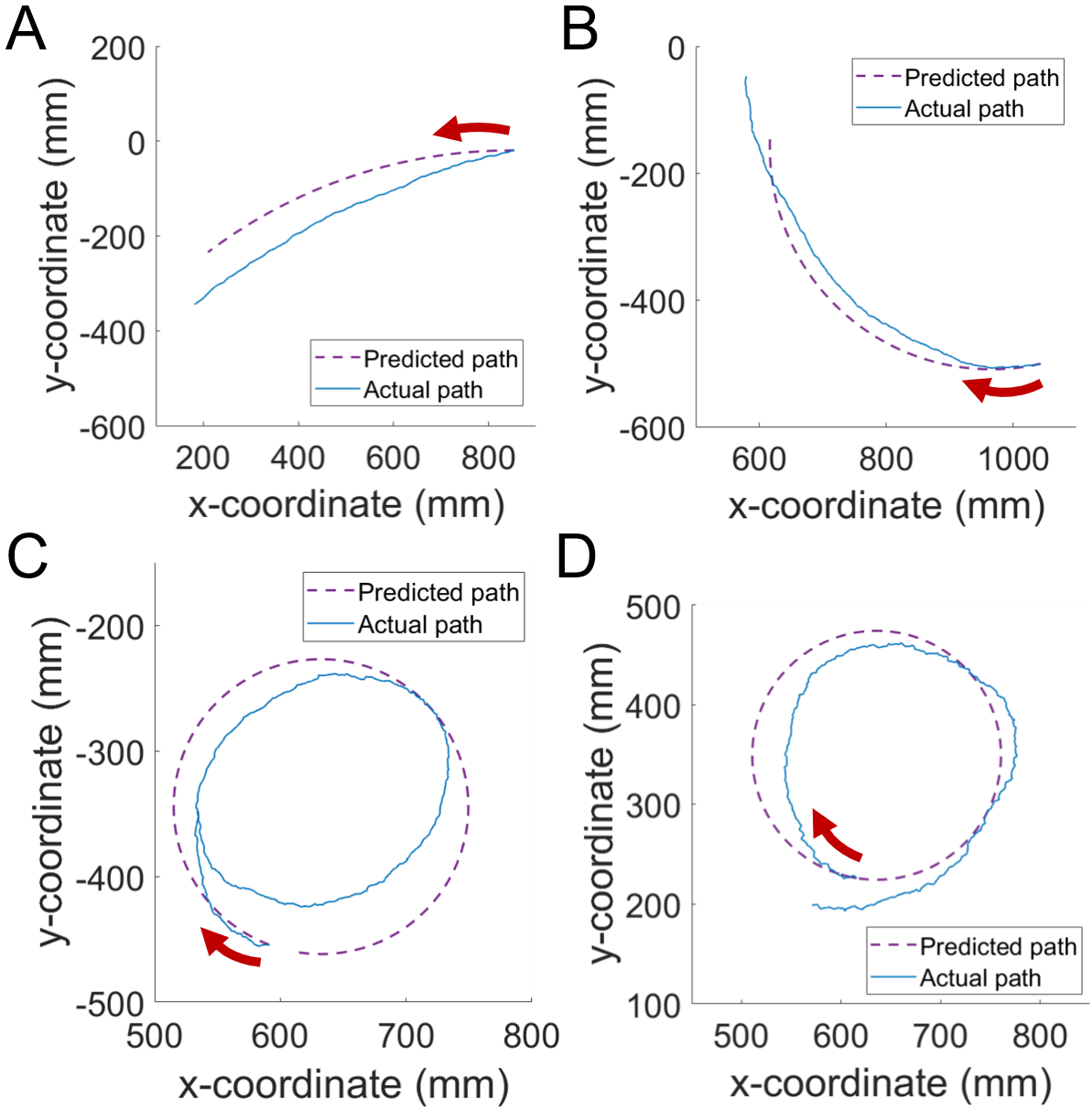}
    \caption{Predicted and actual trajectory of the WHERE-Bot. After fitting the functional relations of motion parameters and structural parameter $\gamma$, trajectory at (A) $\gamma=30\rm\,deg$; (B) $\gamma=90\rm\,deg$; (C) $\gamma=150\rm\,deg$; and (D) $\gamma=210\rm\,deg$ can be predicted as discrete points by Equation \ref{eqn-7}. All discrete points on one trajectory are connected by line segments.}

    \label{figPredict}

\end{figure}
The results are presented in Fig.~\ref{figResults}. Specifically, Fig.~\ref{figResults}(A) shows the trajectories from three experiments at $\gamma=180\rm\,deg$, along with the fitting circle derived from the position data and the standard deviation of the radius. The actual trajectories do not conform to standard circles due to oscillation and complex friction behaviors caused by excessive clearance between the spring coils and the hub teeth. This design facilitates smooth transmission and prevents excessive strain that could damage the spring. Fig.~\ref{figResults}(C) shows that the robot can orbit both clockwise and counterclockwise, depending on the value of $\gamma$. When $\Delta\phi$ approaches zero, the robot theoretically moves along a straight line (Fig.~\ref{figResults}(B)). However, in practice, friction and oscillation interfere with the motion, resulting in a significant error in the measured radius at $\gamma=180\rm\,deg$. As $\gamma$ increases, the robot’s moving distance in each period tends to decrease (Fig.~\ref{figResults}(D)). The moving direction (relative to the reference direction) is a monotonically increasing function of $\gamma$ (Fig.~\ref{figResults}(E)). When the steering module is positioned opposite to the spiral servo, friction $f_1$ and $f_2$ counteract each other to the greatest extent, causing the motion of the spiral servo to decelerate, leading to the longest motion period (Fig.~\ref{figResults}(F)).

To predict the robot’s trajectories under different values of structural parameter $\gamma$, we fitted the relations between motion parameters and $\gamma$ using common functions. The fitting curves are shown in Fig.~\ref{figResults}(B-F). With this data-based model, the robot’s discrete positions between periods can be predicted using the following iterative relation:
\begin{equation}\label{eqn-7} 
\textbf{\textit{p}}(k+1)=\textbf{\textit{p}}(k)+\Delta x[\textbf{R}(\Delta\phi)]^k\textbf{R}(\beta)\textbf{\textit{e}}_{r,0},
\end{equation}
where $\textbf{\textit{p}}(k)$ is the position of the robot's center after $k$ motion periods, $\textbf{R}(\Delta\phi)$ is the rotation matrix for $\Delta\phi, \textbf{R}(\beta)$ is the rotation matrix for $\beta$, and $\textbf{\textit{e}}_{r,0}$ is the reference direction of the robot at the initial position $\textbf{\textit{p}}(0)$. The maximum number of periods, which also represents the maximum iteration count, is calculated by dividing the total motion time by the period obtained from the fitting curves.

Given $\textbf{\textit{p}}(0)$ and $\textbf{\textit{e}}_{r,0}$ for different values of $\gamma$ (90, 150, 210, and 270 degrees), the predicted and actual trajectories are shown in Fig.~\ref{figPredict}(A-D). The predicted trajectories roughly resemble the actual trajectories. Friction and oscillation during motions are the primary factors contributing to the deviations from standard circular trajectories.

\subsection{Trajectory Control}

\begin{figure}
    \centering
    
    \includegraphics[width=0.48\textwidth]{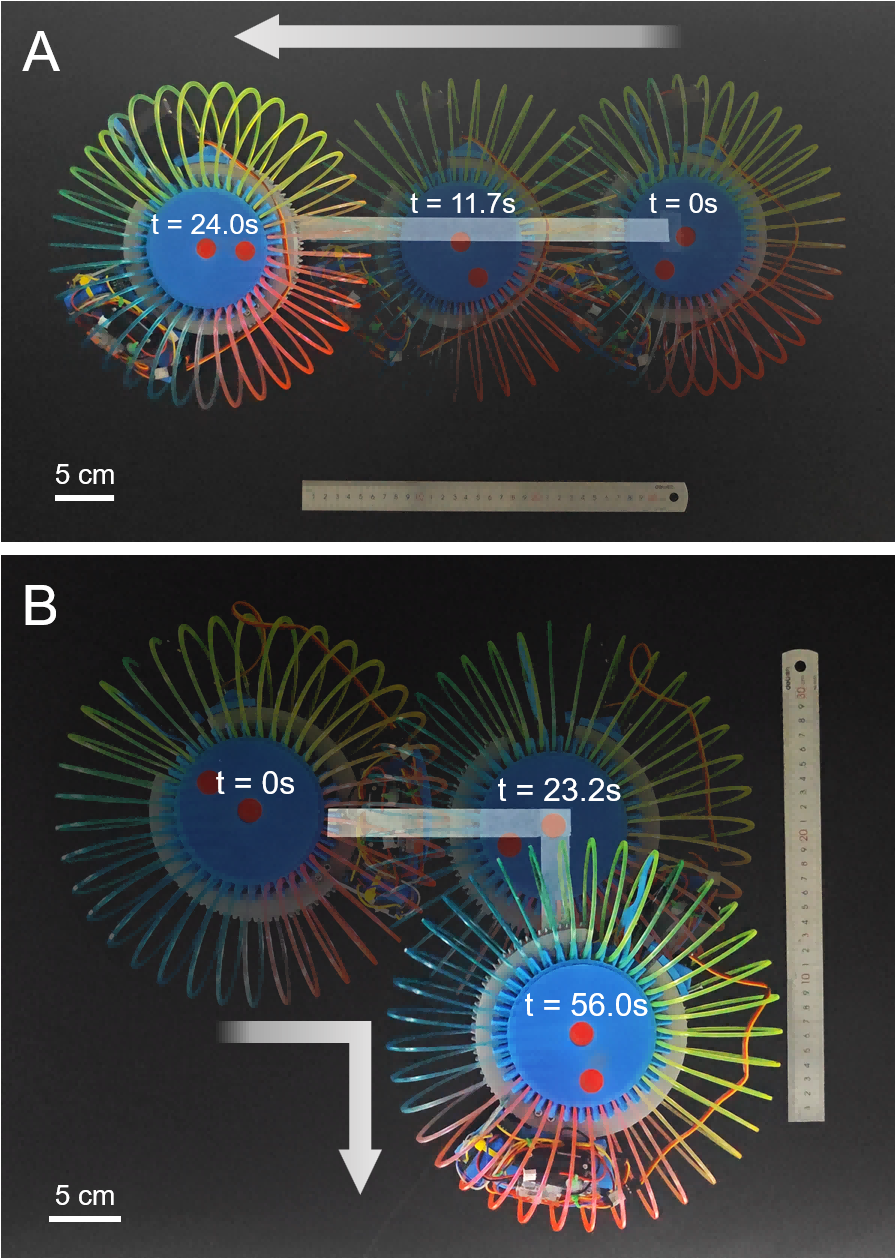}
    \caption{Trajectory control of the WHERE-Bot. The robot's trajectory was observed visually and controlled manually. (A) Moving along a straight path. (B) Moving straight and changing direction.}

        \label{fig:9}

\end{figure}

Without intervention, the robot's motion exhibited a circular trajectory. We attempted to have the WHERE-Bot move along a straight line and make a 90-degree right turn to demonstrate the controllability of its trajectory. During these two tasks, whenever we observed that the moving direction of the WHERE-Bot deviated from the target trajectory marked by white lines, we manually controlled the rotation of the steering servo by a specific angle. This action altered the mass distribution of the robot, thereby changing its moving direction and ensuring that it could follow the target trajectory, as shown in Fig.~\ref{fig:9}.

\subsection{Environment Interaction}

\begin{figure}
    \centering
    
    \includegraphics[width=0.48\textwidth]{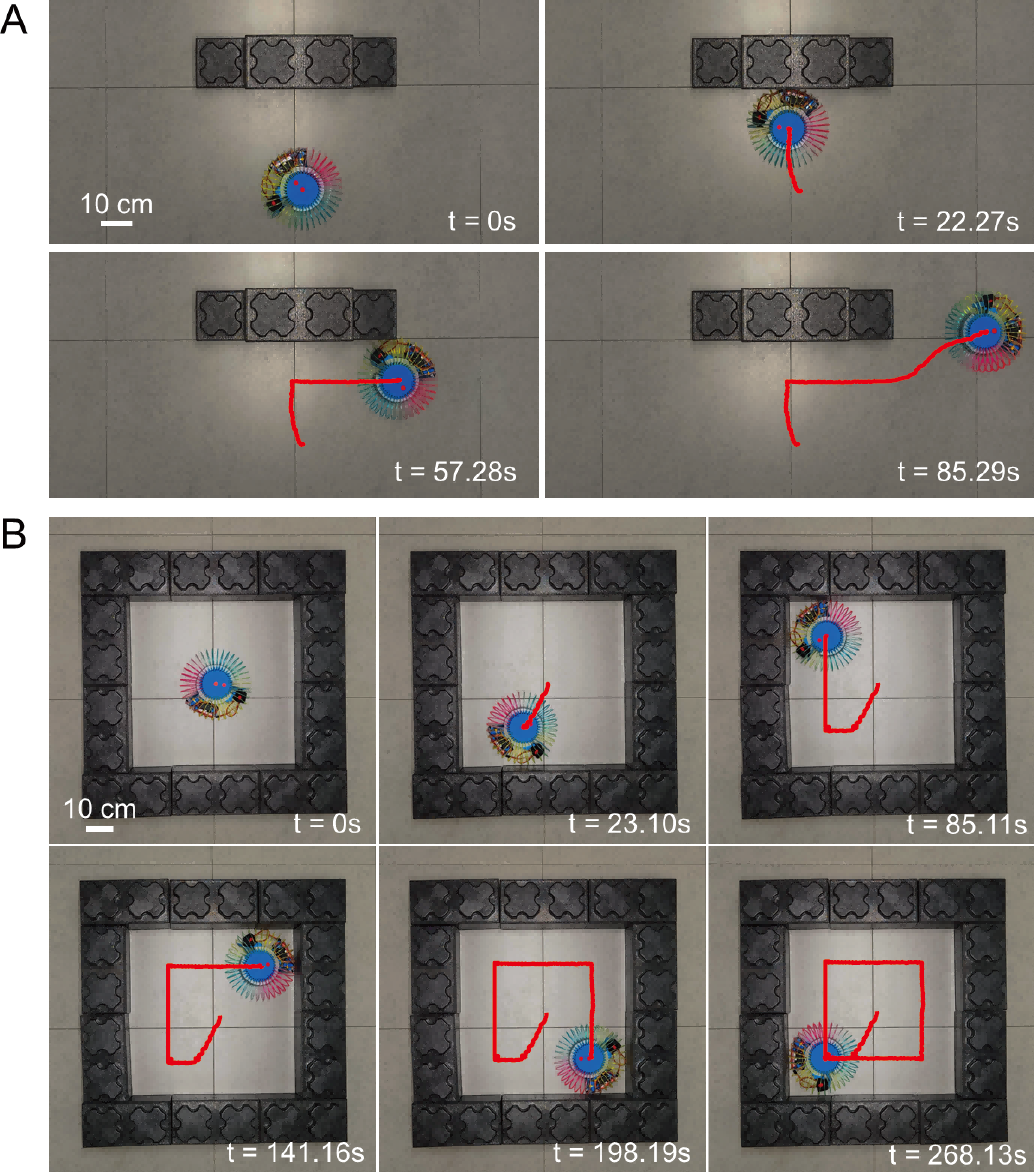}
    \caption{Environment exploration and interaction without sensor and trajectory control. (A) Automated obstacle avoidance behavior. (B) Exploratory behavior within a $60cm\times60cm$ square boundary of the environment. }

    \label{fig:10}

\end{figure}

We tested the robot’s interaction with the environment
without sensor and trajectory control. The robot was placed
on a rigid surface to observe its interaction with environmental boundaries. When the robot encountered an obstacle
during locomotion, its velocity perpendicular to the contact
surface ceased. Additionally, it continued rotating and slid along
the contact surface. Therefore, the robot finally moved away from the
obstacle using the lateral sliding friction (Fig.~\ref{fig:10}(A)).

Subsequently, we positioned the robot within a square boundary environment and observed its spontaneous exploration of the perimeter. With the robot's orbit radius sufficiently large, it made contact with one of the walls and began to crawl along it in a clockwise direction while simultaneously rotating in the same direction. Upon reaching a corner of the boundary, the robot disengaged from the wall which it was originally in contact with and initiated movement along the new wall, as shown in Fig.~\ref{fig:10}(B) and the supplement video. After a certain period of time($268.13s$), the robot was capable of navigating along the $60\rm\,cm\times60\rm\,cm$ square boundary.

\section{Conclusion}
We propose the WHERE-Bot, a wheel-less helical-ring everting mobile robot featuring locomotion pre-defined by mass distribution and reprogrammable trajectories. This robot can navigate through unstructured environments without sensors for detecting obstacles or boundaries, demonstrating potential for boundary exploration.

However, the current design considers a large clearance between the spring coils and hub teeth, which is intended to prevent excessive stress that could break the spring. This results in significant dynamic deformations, complex friction behaviors, and oscillations during the robot's locomotion. Furthermore, the robot is limited to moving on hard surfaces with low friction coefficients to avoid damaging the spring. Future research will focus on fabricating springs with greater structural stiffness and toughness and lower surface frictions to enhance both transmission and motion consistency. By improving the mechanical properties of the spring, the robot's locomotion adaptability to various terrains can be further investigated.

\addtolength{\textheight}{-12cm}   





\bibliography{reference}

\end{document}